\documentclass[10pt,twocolumn, letterpaper]{article}
\usepackage[pagenumbers]{cvpr}

\usepackage{cite}
\usepackage{url}
\usepackage{graphicx}
\usepackage{multirow}
\usepackage{comment}
\usepackage{bm}
\usepackage{pifont}
\usepackage{chngcntr}
\usepackage{makecell}
\usepackage{booktabs}
\usepackage{xcolor}
\usepackage{xspace}
\usepackage{colortbl}
\usepackage{float}
\usepackage{paralist}
\usepackage{wrapfig}
\usepackage{amsmath}
\usepackage{amssymb}

\usepackage[pagebackref=true,breaklinks=true,colorlinks,bookmarks=false,citecolor=blue,linkcolor=blue,urlcolor=blue]{hyperref}
\usepackage[capitalize]{cleveref}

\newcommand{\ch}{\checkmark}
\definecolor{color3}{rgb}{0.95,0.95,0.95}

\newcommand{\txt}[1]{{\texttt{#1}}}

\def\xnet{Burstormer\xspace}

\crefname{section}{Sec.}{Secs.}
\Crefname{section}{Section}{Sections}
\Crefname{table}{Table}{Tables}
\crefname{table}{Tab.}{Tabs.}

\begin{document}

\title{Burstormer: Burst Image Restoration and Enhancement Transformer}

\author{Akshay Dudhane$^1$ \quad Syed Waqas Zamir$^2$ \quad Salman Khan$^{1,3}$ \quad \\
Fahad Shahbaz Khan$^{1,4}$ \quad Ming-Hsuan Yang$^{5,6,7}$ \\
$^1$Mohamed bin Zayed University of AI \hspace{1.5mm} $^2$Inception Institute of AI \hspace{1.5mm} $^3$Australian National University \\
 $^4$Link\"{o}ping University \hspace{1.5mm}  $^5$University of California, Merced \hspace{1.5mm} $^6$Yonsei University \hspace{1.5mm} $^7$Google Research}

\maketitle

\begin{abstract}\vspace{-0.8em}
    On a shutter press, modern handheld cameras capture multiple images in rapid succession and merge them to generate a single image. 
    However, individual frames in a burst are misaligned due to inevitable motions and contain multiple degradations. 
    The challenge is to properly align the successive image shots and merge their complimentary information to achieve high-quality outputs.
    
    Towards this direction, we propose \xnet: a novel transformer-based architecture for burst image restoration and enhancement. 
    In comparison to existing works, our approach exploits multi-scale local and non-local features to achieve improved alignment and feature fusion. 
    Our key idea is to enable  inter-frame communication in the burst neighborhoods for information aggregation and progressive fusion while modeling the burst-wide context. 
    However, the input burst frames need to be properly aligned before fusing their information. 
    Therefore, we propose an {enhanced deformable alignment} module for aligning burst features with regards to the reference frame. 
    
    Unlike existing methods, the proposed alignment module not only aligns burst features but also exchanges feature information and maintains focused communication with the reference frame through the proposed {reference-based feature enrichment} mechanism, which facilitates handling complex motions. 
    After multi-level alignment and enrichment, we re-emphasize on inter-frame communication within burst using a {cyclic burst sampling} module. 
    Finally, the inter-frame information is aggregated using the proposed {burst feature fusion} module followed by progressive upsampling.
    Our \xnet outperforms state-of-the-art methods on burst super-resolution, burst denoising and burst low-light enhancement. Our codes and pre-trained models are available at \url{https://github.com/akshaydudhane16/Burstormer}.
\end{abstract}\vspace{-0.25em}

\begin{figure}[t]
    \centering
    \includegraphics[width=1\linewidth]{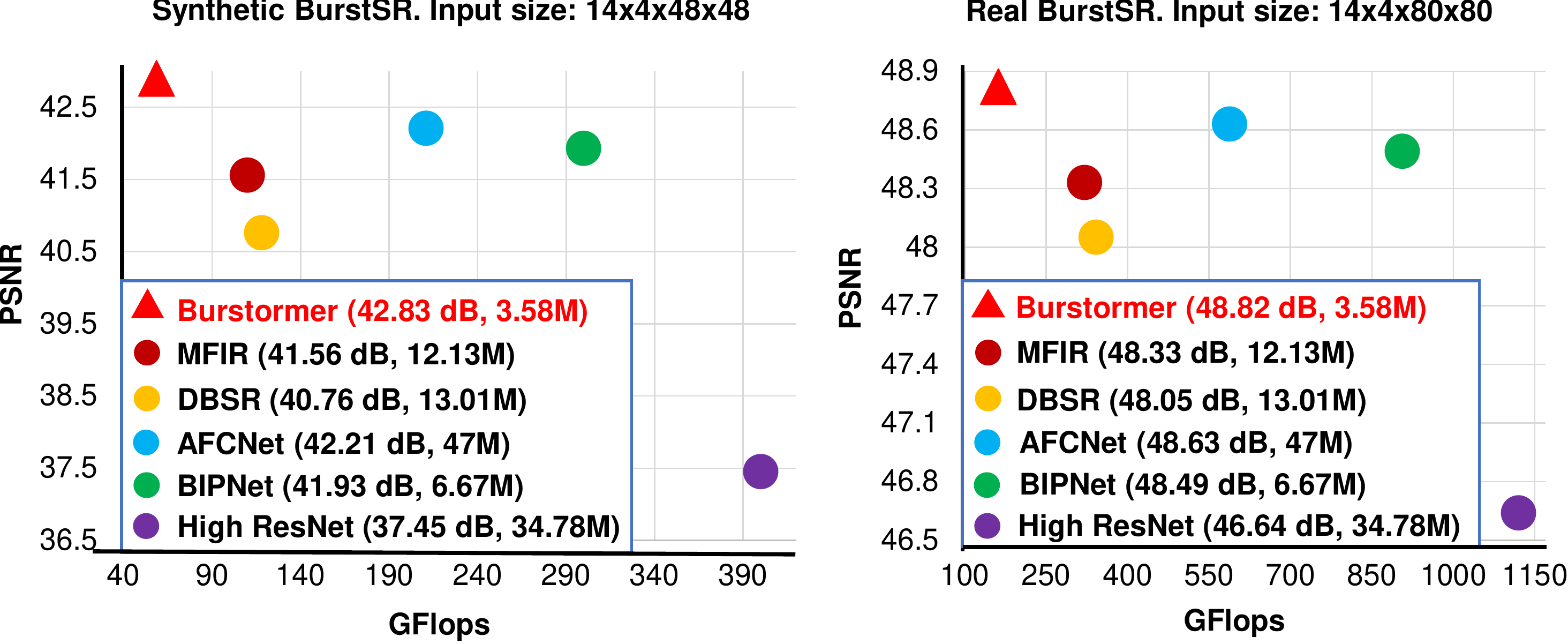}
    \caption{Burst super-resolution results (Tab.~\ref{tab: burstSR}) vs. efficiency (GFlops). \xnet advances state-of-the-art, while being compute efficient and  light-weight.} 
    \label{fig: par_psnr}
\end{figure}

\section{Introduction}
    In recent years, smartphone industry has witnessed a rampant growth on account of the fueling demand of smartphones in  day-to-day life.
    While the image quality of smartphone cameras is rapidly improving, there are several barriers that hinder in attaining DSLR-like images. 
    For instance, the physical space available in handheld devices restricts manufacturers from employing high-quality bulky camera modules.
    Most smartphone cameras use small-sized lens, aperture, and sensor, thereby generating images with limited spatial resolution, low dynamic range, and often with noise and color distortions especially in low-light conditions. 
    These problems have shifted the focus nowadays in developing computational photography (software) solutions for mitigating the hardware limitations and to approach the image quality of DSLRs.

    One emerging approach to achieve high-quality results from a smartphone camera is to take advantage of burst shots consisting of multiple captures of the same scene.
    The burst image processing approaches aim to recover the high-quality image by merging the complementary information in multiple frames.
    Recent works~\cite{dudhane2022burst, bhat2021deep, bhat2021deep1} have validated the potential of burst processing techniques in reconstructing rich details that cannot be recovered from a single image. 
    However, these computationally expensive approaches are often unable to effectively deal with the inherent sub-pixel shifts among multiple frames arising due to camera and/or object movements.
    This sub-pixel misalignment often causes blurring and ghosting artifacts in the final image.
    To tackle alignment issues, existing methods employ complex explicit feature alignment~\cite{bhat2021deep} and deformable convolutions~\cite{dudhane2022burst}.
    However, these approaches target only the local features at a single level, while the use of global information together with multi-scale features has not been extensively explored.
    Additionally, while aggregating multi-frame features, existing approaches either employ late fusion strategy~\cite{bhat2021deep, bhat2021deep1} or rigid fusion mechanism (in terms of number of frames)~\cite{dudhane2022burst}.
    The former one limits the flexible inter-frame communication, while the later one limits the adaptive multi-frame processing.   
    
    In this work, we propose \xnet for burst image processing, which incorporates multi-level local-global burst feature alignment and adaptive burst feature aggregation.
    In contrast to previous works~\cite{bhat2021deep, bhat2021deep1} that employ bulky pre-trained networks for explicit feature alignment, we present a novel {enhanced deformable alignment (EDA) module that handles misalignment issues implicitly.}
    Overall, the EDA module reduces noise and extracts both local and non-local features with a transformer-based attention and performs multi-scale burst feature alignment and feature enrichment which is not the case with the recent BIPNet \cite{dudhane2022burst}.    
    
    Unlike existing approaches~\cite{dudhane2022burst,bhat2021deep, bhat2021deep1} which allow a one go interaction with the reference frame during alignment process, we add a new reference-based feature enrichment (RBFE) mechanism in EDA to allow a more extensive interaction with the reference frame.
    This helps in effectively aligning and refining burst features even in complex misalignment cases where the simple alignment approaches would not suffice.
    In the image reconstruction stage we progressively perform feature consolidation and upsampling, while having access to the multi-frame feature information at all time.
    This is achieved with our no-reference feature enrichment (NRFE) module. NRFE initially generates burst neighborhoods with the proposed cyclic burst sampling (CBS) mechanism that are then aggregated with our burst feature fusion (BFF) unit.
    Unlike, the existing pseudo bursts~\cite{dudhane2022burst}, the proposed burst neighborhood mechanism is flexible and enables inter-frame communication with significantly less computational cost.

    \noindent The key highlights of our work are outlined below: 
    \begin{compactitem}
        \item Our \xnet is a novel Transformer based design for burst-image restoration and enhancement that leverages multi-scale local and non-local features for improved alignment and feature fusion. Its flexible design allows processing bursts of variable sizes.
        \item We propose an enhanced deformable alignment module which is based on multi-scale hierarchical design to effectively denoise and align burst features. Apart from aligning burst features it also refines and consolidates the complimentary burst features with the proposed reference-based feature enrichment module. 
        \item We propose no-reference feature enrichment module to progressively aggregate and upsample the burst features with less computational overhead. To enable inter-frame interactions, it generates burst neighborhoods through the proposed cyclic burst sampling mechanism followed by the burst feature fusion.
    \end{compactitem}
    Our \xnet sets new state-of-the-art on several real and synthetic benchmark datasets for the task of burst super-resolution, burst low-light enhancement, and burst denoising. Compared to existing approaches, \xnet is more accurate, light-weight and faster; see Fig.~\ref{fig: par_psnr}. Further, we provide detailed ablation studies to demonstrate the effectiveness of our design choices.

\section{Related Work}
\label{sec: related work}
    \noindent \textbf{Multi-Frame Super-Resolution.}
        Unlike single image super-resolution, multi-frame super-resolution (MFSR) approaches are required to additionally deal with the sub-pixel misalignments among burst frames caused by camera and object motions.
        While being computationally efficient, the pioneering MFSR algorithm~\cite{tsai} processes burst frames in frequency domain, often producing images with noticeable artifacts. 
        To obtain better super-resolved results, other methods operate in the spatial domain~\cite{spatial,irani}, exploit image priors~\cite{stark1989high}, use iterative back-projection~\cite{peleg}, or maximum a posteriori framework~\cite{map1}.
        However, all these approaches assume that the image formation model, and motion among input frames can be estimated reliably. Successive works addressed this constraint with the joint estimation of the unknown parameters~\cite{u1, u2}. 
        To deal with noise and complex motion, the MSFR algorithm of \cite{wronski} employs non-parametric kernel regression and locally adaptive detail enhancement model.  

        The DBSR algorithm~\cite{bhat2021deep} addresses the MFSR problem by applying the explicit feature alignment and attention-centric fusion mechanisms.
        However, their image warping technique and explicit motion estimation may find difficult in handling scenes with fast moving objects.
        The EBSR~\cite{luo2021ebsr} builds on prior PCD alignment techniques~\cite{wang2019edvr} by aligning enhanced features specifically for the burst SR task. In addition, the BSRT~\cite{luo2022bsrt} employs a combination of optical flow and deformable convolution for feature alignment and utilizes a Swin Transformer~\cite{liu2021swin} for feature extraction.
        More recently, BIPNet~\cite{dudhane2022burst} was introduced to process noisy raw bursts using implicit feature alignment and pseudo-burst generation. Building on BIPNet, AFCNet~\cite{mehta2022adaptive} incorporates existing Restormer~\cite{zamir2021restormer} to improve feature extraction for burst SR tasks.
        Despite having an effective inter-frame communication, their approach is rigid to using certain number of burst frames during alignment and fusion.
    
    \vspace{0.4em}
    \noindent \textbf{Multi-Frame Denoising.}
        Aside from aforementioned MFSR approaches, several multi-frame methods have been developed to perform denoising~\cite{Dabov2007VideoDB, Maggioni2011VideoDU, Maggioni2012VideoDD, hasinoff2016burst}.
        The algorithm of ~\cite{Tico2008MultiframeID} leverages visually similar image block within and across frames to obtain denoised results.
        Other works \cite{Dabov2007VideoDB, Maggioni2012VideoDD} extend the state-of-the-art single image denoising technique BM3D~\cite{Dabov2007VideoDB} to videos. 
        The method of \cite{liu1} yields favorable denoising results by employing a novel homography flow alignment technique with consistent pixel compositing operator.
        In the work of \cite{Godard2018DeepBD}, the authors extend single-image denoising network to multi-frame task via recurrent deep convolutional neural network.
        The kernel prediction network \cite{Mildenhall2018BurstDW} generates per-pixel kernels for fusing multiple-frames.
        RViDeNet~\cite{yue2020supervised} uses deformable convolutions to perform explicit frame alignment in order to provide improved denoising results.
        The re-parametrization approach of MFIR~\cite{bhat2021deep1} learns image formation model in deep feature space for the multi-frame denoising.
        BIPNet~\cite{dudhane2022burst} presents a novel pseudo-burst feature fusion approach to perform denoising on burst frames.

    \vspace{0.4em}
    \noindent \textbf{Multi-Frame Low-light Image Enhancement.}
        In low-light conditions, smartphone cameras often yield noisy and color-distorted images due to their small aperture and sensor pixel cavities. 
        \cite{chen2018learning} collect a multi-frame dataset for low-light image enhancement, and present a data-driven approach to learn camera imaging pipeline in order to map under-exposed RAW images directly to well-lit sRGB images. 
        The quality of output image is further improved with the perceptual loss presented by~\cite{zamir2021learning}.
        The works of \cite{maharjan2019improving} and \cite{zhao2019end}, respectively, use residual learning framework and recurrent convolution network to obtain enhanced images from multiple degraded low-lit input frames. 
        The two-stage approach of \cite{karadeniz2020burst} employs one subnet for explicitly denoising multiple frames followed by the second subnet to provide us with the enhanced image. 
        Along with super-resolution and denoising, BIPNet~\cite{dudhane2022burst} is also capable of performing multi-frame low-light image enhancement.
        Unlike the existing multi-frame approaches, our \xnet aligns burst features at multiple-scales and enables flexible inter-frame communication without much computational overhead. It also incorporates progressive feature merging to obtain high-quality images. 
        
        \begin{figure*}[t]
            \centering
            \includegraphics[width=0.98\linewidth]{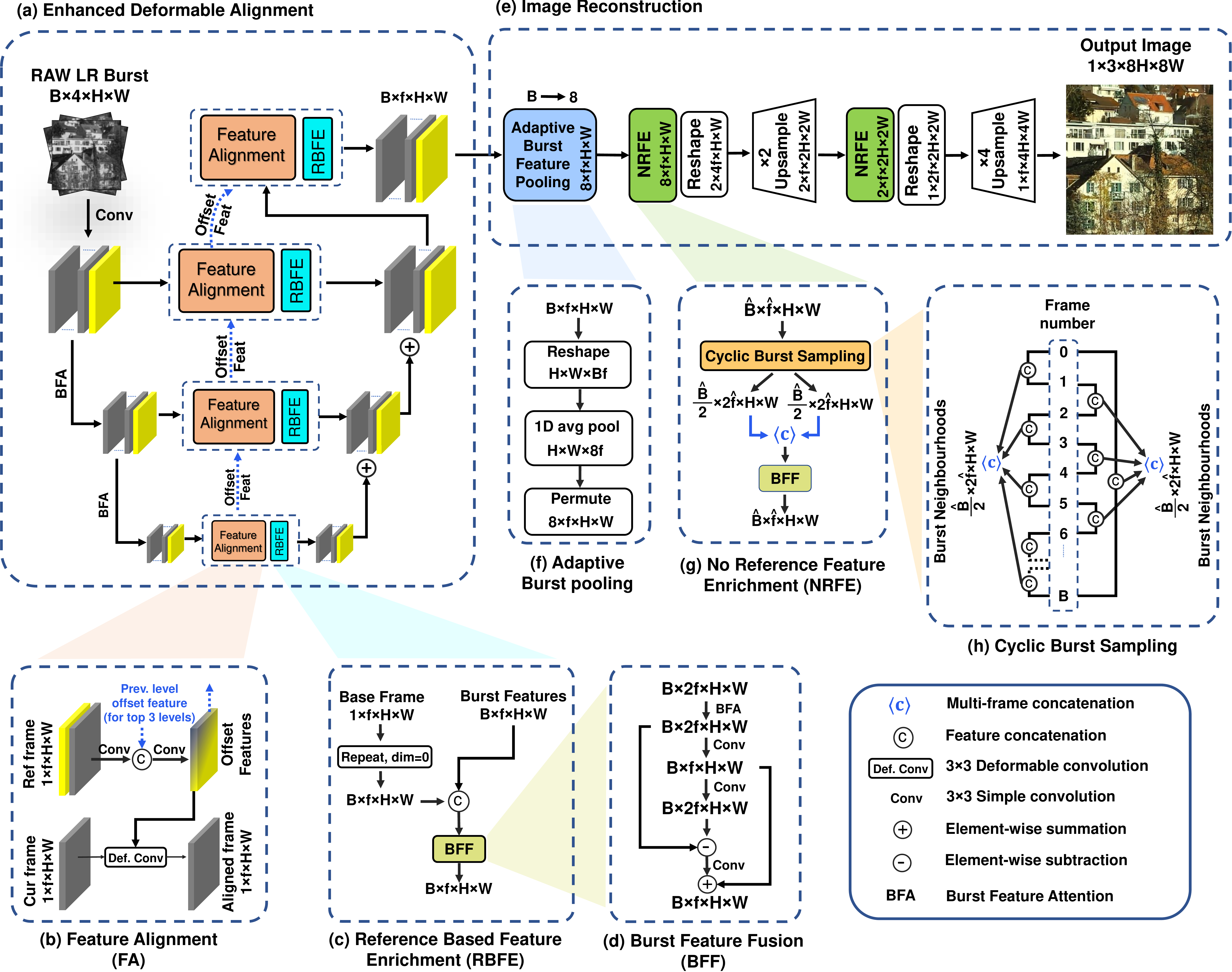}
            \caption{Overall pipeline of the proposed \xnet for burst image processing.
            \xnet takes as input a RAW burst of degraded images and outputs a clean high-quality sRGB image.
            It has two main parts: enhanced deformable alignment (EDA), and the image reconstruction. 
            EDA, labeled as \txt{(a)}, is a multi-scale hierarchical module that, at each level, first extracts noise-free local and non-local features with the burst feature attention (BFA), performs feature alignment \txt{(b)}, and finally refines and consolidates features with an additional interaction with the base frame via \txt{(c)} the proposed reference-based feature enrichment (RBFE) module. RBFE further employs \txt{(d)} the burst feature fusion (BFF) unit for merging features by using the back-projection and squeeze-excitation mechanisms.
            The aligned burst of features are then passed to the image reconstruction stage \txt{(e)}. Here \txt{(f)} the adaptive burst pooling module transforms the input burst size (B frames) to constant 8 frames through an average pooling operator. Finally, \txt{(g)} the no-reference feature enrichment (NRFE) module progressively aggregates, and upsamples the burst features to generate the final HR image.}
            \label{fig: 1}
        \end{figure*}
    
\section{Proposed Burst Image Processing Pipeline}
    Burst sequences are usually acquired with handheld devices. The spatial and color misalignments among burst frames are unavoidable due to hand-tremor and camera/object motions. These issues negatively affect the overall performance of the burst processing approaches. In this work, our goal is to effectively \emph{align} and progressively \emph{merge} the desired information from multiple degraded frames to reconstruct a high-quality composite image.   
    To this end, we propose \xnet, a novel unified model for multi-frame processing where different modules jointly operate to perform feature denoising, alignment, fusion, and upsampling tasks. 
    Here, we describe our method for the task of burst super-resolution, nevertheless, it is applicable to different burst restoration tasks such as burst denoising and burst enhancement (see experiments Sec.~\ref{sec:experiments}).

\vspace{0.4em}
 \noindent   \textbf{Overall Pipeline.} Fig. \ref{fig: 1} shows the overall pipeline of the proposed \xnet.
    \emph{First,} the RAW input burst is passed through the proposed enhanced deformable alignment (EDA) module which extract noise-free deep features that are aligned and refined with respect to the reference frame features.
    \emph{Second,} an image reconstruction module is employed that takes as input the burst of aligned features and progressively merges them using the proposed no reference feature enrichment (NRFE) module. To obtain the super-resolved image, the upsampling operation is immediately applied after each NRFE module in the reconstruction stage.  
    Next, we describe each stage of  our approach. 
    \subsection{Enhanced Deformable Alignment}
        In burst processing, effective alignment of mismatched frames is essential as in any error arising in this stage will propagate to later stages, subsequently making the reconstruction task difficult. 
        Existing methods perform image alignment either explicitly~\cite{bhat2021deep, bhat2021deep1}, or implicitly~\cite{dudhane2022burst}. While, these techniques are suitable to correcting mild pixel displacements among frames, they might not adequately handle fast moving objects. 
        In \xnet, we propose enhanced deformable alignment (EDA) which employs a multi-scale design as shown in Fig.~\ref{fig: 1}\textcolor{blue}{(a)}. Since sub-pixel shifts among frames are naturally reduced at low-spatial resolution, using the multi-level hierarchical architecture provides us with more robust alignment. Therefore, EDA starts feature alignment from the lowest level (3$^{rd}$ in this paper) and progressively passes offsets to upper high-resolution levels to help with the alignment process.
        Furthermore, at each level, the aligned features are passed through the proposed reference-based feature enrichment (RBFE) module to fix remaining misalignment issues in burst frames by interacting them again with the reference frame.
        EDA has two key components: \textbf{(1)} Feature alignment, and \textbf{(2)} Reference-based feature enrichment. 

      \vspace{0.4em}
      \noindent \textbf{Feature alignment.}
            Burst images are often contaminated with random noise that impedes in finding the dense correspondences among frames. Therefore, before performing alignment operation, we extract noise-free burst features by using burst feature attention (BFA) module which is built upon the existing transformer block~\cite{zamir2021restormer}. Unlike in other approaches~\cite{bhat2021deep,bhat2021deep1,dudhane2022burst}, the BFA module encodes local and non-local context using MDTA block~\cite{zamir2021restormer} and controls feature transformation through the GDFN~\cite{zamir2021restormer} block. Furthermore, unlike existing attention techniques~\cite{vaswani2017attention, liu2021swin, thawakar2022video}, BFA module is efficient enough to be applied to high-resolution images.
            The denoised features from BFA are passed further for alignment.
            Figure~\ref{fig: 1}\textcolor{blue}{(b)} shows the feature alignment (FA) module that utilizes a modulated deformable convolution~\cite{zhu2019deformable} to align features of each burst frame to those of the reference frame.
            Let, $\left\{ {{\bm{g}^b}}: b \in  [1, \ldots, B]\right\} \in \mathbb{R}^{B \times f \times H \times W}$ denotes the burst features obtained after BFA module, where $B$ denotes number of burst frames, $f$ is the number of feature channels, and $H$$\times$$W$ is the spatial size. We align the features of the current frame $\bm{g}^{b}$ with the reference frame\footnote{We consider the first burst image to be the reference frame.} $\bm{g}^{b_r}$. Feature alignment module processes $\bm{g}^b$ and $\bm{g}^{b_r}$ via an offset convolution layer and outputs the offset $\Delta {n}$ and modulation scalar $\Delta a$ values for $\bm{g}^b$. In Fig. \ref{fig: 1}(a), for simplicity, only offset $\Delta {n}$ is shown.
            The aligned features ${\bm{\bar g}}^b$ are computed as:
            \begin{equation}
                {{\bm{\bar g}}^b} = {W_{\text{def}}}\left( {{\bm{g}^b},\; \{\Delta {n},\;\Delta a\}} \right),\;
                \\
                \{\Delta n, \Delta a\} = {W_{\text{off}}}\left( {{\bm{g}^b},\;{\bm{g}^{{b_r}}}} \right),
            \end{equation}
            where, $W_{\text{def}}(\cdot)$ and $W_{\text{off}}(\cdot)$ represent the deformable and offset convolutions, respectively. Specifically, every position $n$ on the aligned feature map ${\bm{\bar g}}^b$ is calculated as:
            \begin{equation}
                {{\bm{\bar g}}^b}_n = \sum\limits_{i=1}^K {{W_{n_i}^{d}}\,\,\,{\bm{y}_{\left( n + {n_i} + \Delta {n_i} \right)}^b}} \cdot \Delta {a_{n_i}},
            \end{equation}
            where, $K$=9, $\Delta a$ in range $[0, 1]$ for each $n_i \in \left\{ { (-1, 1),  \allowbreak (-1, 0), ..., (1,1)} \right\}$ is a regular $3{\times}3$ kernel grid. The convolution is performed on non-uniform positions $({n_i} + \Delta {n_i})$, where ${n_i}$ may be fractional. To tackle the fractional values, this operation is implemented with the bilinear interpolation. 

       \vspace{0.4em}
       \noindent \textbf{Reference-Based Feature Enrichment.}
            In the presence of complex pixel displacements among frames, simple alignment techniques \cite{bhat2021deep, bhat2021deep1, dudhane2022burst} may not able to align burst features completely. Thus, to fix the remaining minor misalignment issues, we propose the reference-based feature enrichment (RBFE) module, shown in Fig.~\ref{fig: 1}\textcolor{blue}{(c)}. 
            RBFE enables additional interaction of aligned frames features $\bm{\bar g}^b$  with the reference frame features $\bm{g}^{b_r}$ to generate consolidated and refined representations. 
            This interactive feature merging is performed via our burst feature fusion (BFF) unit as illustrated in Fig. \ref{fig: 1}\textcolor{blue}{(d)}. 
            The BFF mechanism is built upon the principles of feature back projection~\cite{haris2018deep} and squeeze-excitation techniques~\cite{hu2018squeeze}.
            Given the concatenated feature maps of the current frame and the reference frame $[\bm{\bar g}^b, \bm{g}^{b_r}] \in{\mathbb{R}^{1 \times 2\text{*}f \times H \times W}}$, BFF applies BFA to generate representations $\bm{g^b_a}$ encoding the local non-local context. Overall, BFF yields fused features $\bm{g^b_f}\in{\mathbb{R}^{1 \times f \times H \times W}}$:
            \begin{equation}
                {\bm{g^b_f}} = {\bm{g^b_s}} + W\left({{\bm{g^b_a}}} - {\bm{g^b_e}} \right),
            \end{equation}
            where $\bm{g^b_s}{=}{W_s}\bm{g^b_a}$ represents squeezed features and $\bm{g^b_e}{=}W_eW_s\bm{g^b_a}$ are the expanded features. $W_s$ and $W_e$ denote simple convolutions to squeeze and expand feature channels. %
            The squeezed features $\bm{g^b_s}$ poses complementary properties of multiple input features. While, $\bm{g^b_e}$ is used to compute the high-frequency residue with the attentive features $\bm{g^b_a}$.
            The aggregation of this high-frequency residual with the squeezed features $\bm{g^b_s}$ helps to learn the feature fusion process implicitly and provides the capability to extract high-frequency complementary information from multiple inputs.
            While illustrated for fusing features of two frames in Fig. \ref{fig: 1}\textcolor{blue}{(d)}, the proposed BFF can be flexibly adapted to any number of inputs.
        
    \subsection{Image Reconstruction}\label{sec:nrbfe_cyclic}
        Figure~\ref{fig: 1}\textcolor{blue}{(e)} illustrates the overall image reconstruction stage.
        To operate on bursts of arbitrary sizes, we propose an adaptive burst feature pooling (ABFP) mechanism that returns the constant burst-size features. As shown in Fig.~\ref{fig: 1}\textcolor{blue}{(f)}, the burst features ($B*f$) are concatenated along channel dimension followed by 1D average pooling operation which adaptively pools out the burst features ($8*f$) as per the requirement.
        Next, the pooled burst feature maps pass through the
        no-reference feature enrichment (NRFE) module, shown in Fig.~\ref{fig: 1}\textcolor{blue}{(g)}.
        The key idea of the proposed NRFE module is to pair immediate neighborhood frames along feature dimension and fuse them using the BFF module.
        However, doing this would limit the inter-frame communication to successive frames only. Therefore, we propose cyclic burst sampling (CBS) that gathers the neighborhood frames in zigzag manner (called here as burst neighborhoods) such that reference frame could interact with the last frame as well via intermediate frames; see Fig.~\ref{fig: 1}\textcolor{blue}{(h)}.
        This cyclic scheme of sampling the burst frames helps in long-range communication without increasing the computational overhead unlike the existing pseudo burst technique~\cite{dudhane2022burst}.
        Next, the sampled neighborhood features are combined along burst dimension and processed with BFF to integrate the useful information available in multiple frames of the burst sequence.

        To upscale the burst features, we adapt pixel-shuffle~\cite{shi2016real} such that the information available in burst frames is shuffled to increase the spatial resolution. 
        This helps in reducing the compute cost and the overall network parameters.    
       
\section{Experiments and Analysis}\label{sec:experiments}
    We evaluate the performance of the proposed \xnet on three different burst image processing tasks: \textbf{(a)} super-resolution (on synthetic and real burst images), \textbf{(b)} low-light image enhancement, and \textbf{(c)} denoising (on grayscale and color data). Additional visual results, ablation experiments, and more details on the network and training settings are provided in the supplementary material.
      
    \vspace{0.4em}
    \noindent \textbf{Implementation Details.}
    We train separate models for different tasks in an end-to-end manner without pre-training any module. 
    We pack the input mosaicked raw burst into 4-channel RGGB format. All burst frames are handled with shared \xnet modules (FA, RBFE, BFF, NRFE) for better parameter efficiency. 
    The following training settings are common to all tasks, whereas task-specific experimental details are provided in their corresponding sections.
    The EDA module of \xnet is a 3-level encoder-decoder, where each level employs 1 FA (containing single deformable conv. layer) and 1 RBFE module. The BFF unit both in RBFE and NRFE consists of 1 BFA module. Each BFA module consists of 1 multi-dconv head transposed attention (MDTA) and 1 gated-Dconv feed-forward network (GDFN)~\cite{zamir2021restormer}. In the image reconstruction stage, we use 2 NRFE modules.
    We train models with $L_1$ loss and Adam optimizer with the initial learning rate $1e^{-4}$ that is gradually reduced to $1e^{-6}$ with the cosine annealing scheduler~\cite{loshchilov2016sgdr} on four RTX6000 GPUs. 
    Random horizontal and vertical image flipping is used for data augmentation. 
        
    \subsection{Burst Super-resolution}
        We evaluate the proposed \xnet on synthetic as well as on real-world datasets~\cite{bhat2021deep, bhat2021ntire} for the SR scale factor~$\times$4. 
        For comparisons, we consider several burst SR approaches such as DBSR~\cite{bhat2021deep}, LKR~\cite{lecouat2021lucas}, HighResNet~\cite{deudon2020highres}, MFIR~\cite{bhat2021deep1} and BIPNet~\cite{dudhane2022burst}.
        
        \vspace{0.4em}
        \noindent \textbf{Datasets.} \textbf{(1)} SyntheticBurst dataset~\cite{bhat2021deep} contains 46,839 RAW burst sequences for training and 300 for validation. 
        Each sequence consists of 14 LR RAW images (with spatial resolution of 48$\times$48 pixels) that are synthetically generated from a single sRGB image as follows.  
        The given sRGB image is first transformed to RAW space with the inverse camera pipeline~\cite{brooks2019unprocessing}. Next, random rotations and translations are applied to this RAW image to generate the HR burst sequence. The HR burst is finally converted to LR RAW burst sequence by applying the downsampling, Bayer mosaicking, sampling and random noise addition operations.\newline
        \textbf{(2)} BurstSR dataset~\cite{bhat2021deep} has 200 RAW burst sequences, each containing 14 images. The LR images of these sequences are captured with a smartphone camera, whereas their corresponding HR (ground-truth) images are taken with a DSLR camera.
        From 200 full-resolution sequences, the original authors extract 5,405 patches of size 80$\times$80 for training and 882 patches for validation. 

\begin{figure}[t]
                \centering
                \includegraphics[width=1\linewidth]{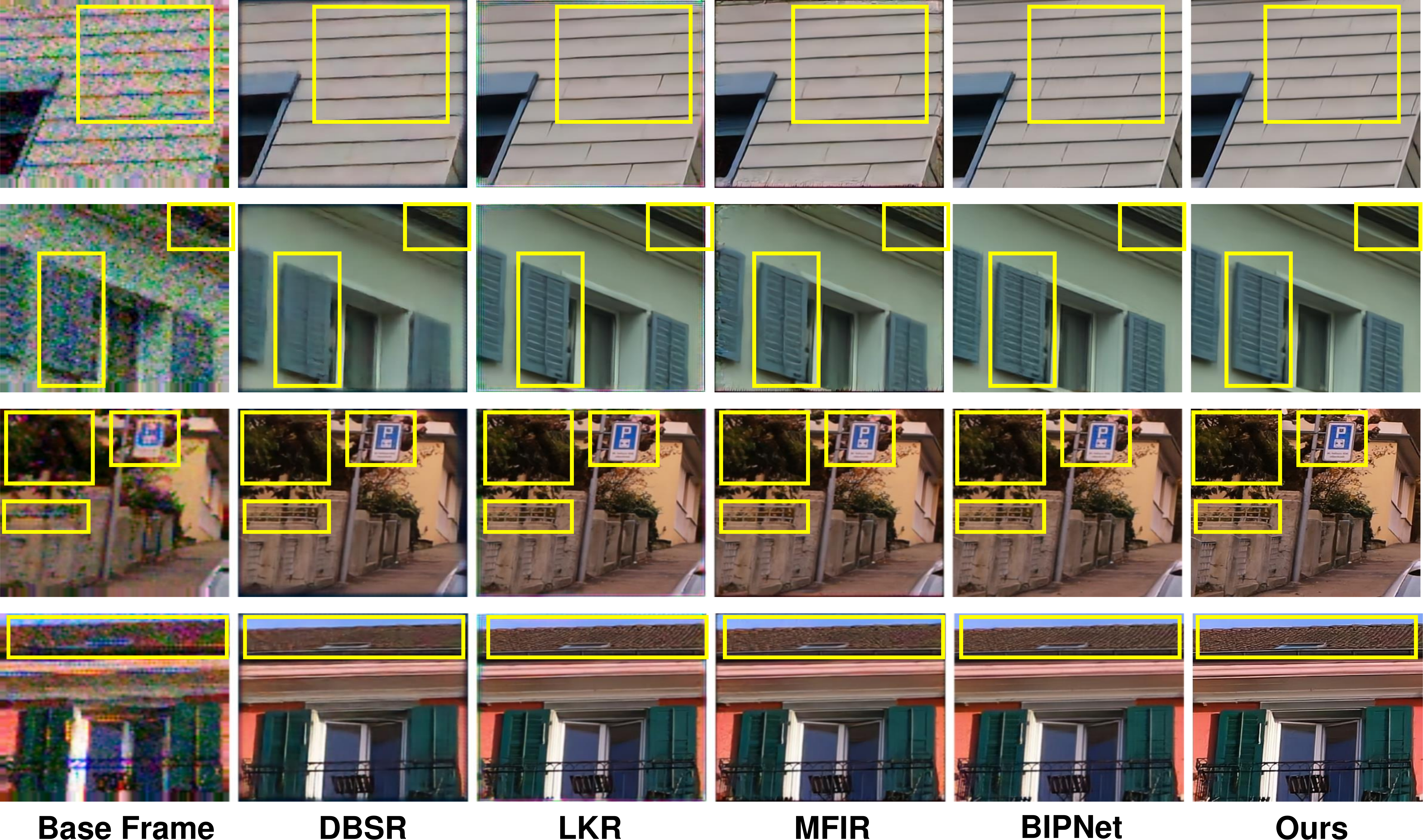}
                \caption{\textbf{Burst super-resolution} ($\times$4) results on SyntheticBurst dataset~\cite{bhat2021deep}. The SR images by our \xnet retain more texture and structural content than the other approaches.}
                \label{fig: syntheticburst}
            \end{figure}

        \vspace{0.4em}
        \noindent \textbf{SR results on synthetic dataset.}
            We train \xnet with batch size 4 for 300 epochs on SyntheticBurst dataset~\cite{bhat2021deep}. 
            Table~\ref{tab: burstSR} shows that our approach significantly advances the state of the art. 
            When compared to the previous best BIPNet~\cite{dudhane2022burst}, our \xnet yields performance gain of 0.9 dB, while having 47$\%$ fewer parameters, 80$\%$ less FLOPs, and runs $2\times$ faster.
            Fig.~\ref{fig: syntheticburst} shows that \xnet-restored images are visually superior with enhanced structural and textural details compared to competing methods. 
            Specifically, the reproductions of DBSR~\cite{bhat2021deep}, LKR~\cite{lecouat2021lucas}, and MFIR~~\cite{bhat2021deep1} contain blotchy textures and color artifacts. 
                 
        \begin{table}[t]
            \centering
            \footnotesize
            \setlength{\tabcolsep}{5pt}
            \caption{\textbf{Burst super-resolution results} on synthetic and real datasets~\cite{bhat2021deep} for factor $4 \times$.}
            \label{tab: burstSR}
            \scalebox{0.9}{
            \begin{tabular}{l@{$\;\,$}|c@{$\;\,$}c@{$\;\,$}c@{$\;\,$}|c@{$\;\,$}c}
                \toprule
                \multirow{2}{4em}{\textbf{Methods}}  & \multicolumn{3}{c}{\textbf{SyntheticBurst}} & \multicolumn{2}{|c}{\textbf{(Real) BurstSR}} \\
                \cmidrule{2-6}
                & \textbf{PSNR $\uparrow$} & \textbf{SSIM $\uparrow$} & \textbf{Time (ms)} & \textbf{PSNR $\uparrow$} & \textbf{SSIM $\uparrow$} \\
                \midrule
                Single Image &  36.17 &   0.91  & 40.0    &  46.29     &  0.982 \\
                HighRes-net \cite{deudon2020highres} & 37.45 & 0.92 & 46.3 & 46.64 & 0.980 \\
                DBSR~\cite{bhat2021deep} & 40.76 & 0.96 & 431 & 48.05 & 0.984 \\
                LKR~\cite{lecouat2021lucas} & 41.45 & 0.95 & -  & - & - \\
                MFIR~\cite{bhat2021deep1} & 41.56 & 0.96 & 420 & 48.33 & 0.985 \\
                BIPNet~\cite{dudhane2022burst} & 41.93 & 0.96 & 130 & 48.49 & 0.985 \\
                AFCNet~\cite{mehta2022adaptive} & 42.21 & 0.96 & 140 & 48.63 & 0.986 \\
                \midrule
                \textbf{\xnet~(Ours)} & \textbf{42.83} & \textbf{0.97} & \textbf{55.0} & \textbf{48.82} & \textbf{0.986} \\
                \bottomrule
            \end{tabular}%
            }
        \end{table}
           
\begin{figure}[t]
                \centering
                \includegraphics[width=1\linewidth]{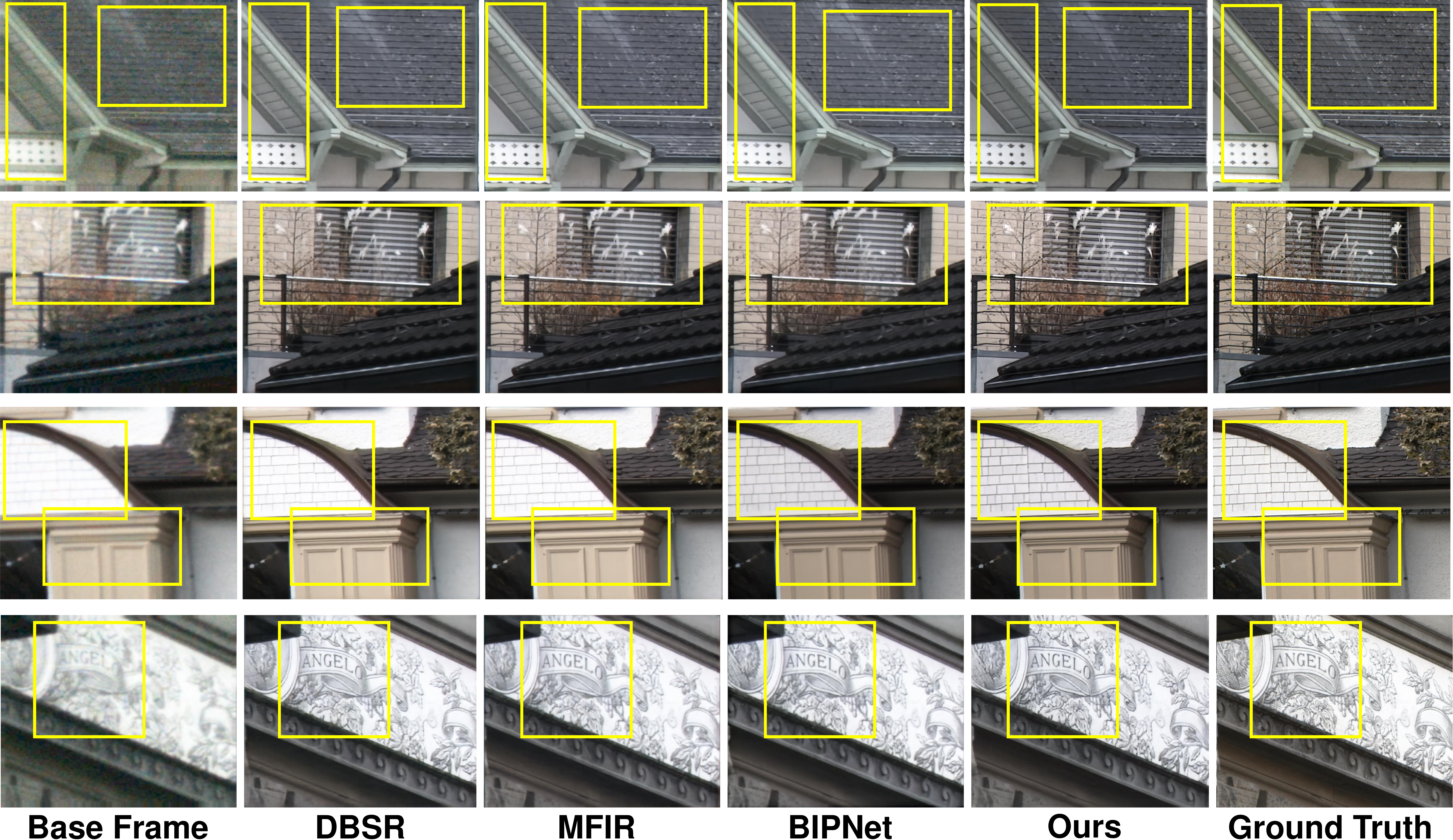}
                \caption{\textbf{Burst super-resolution} ($\times$4) results on BurstSR dataset~\cite{bhat2021deep}. Our results recover better visual details.}
                \label{fig: burstSR}
            \end{figure}

        \vspace{0.4em}
        \noindent \textbf{SR results on real dataset.}
            In BurstSR dataset~\cite{bhat2021deep}, the LR and HR bursts are slightly misaligned as they are captured with different cameras.
            We address this by training \xnet using aligned $L_1$ loss and evaluating with aligned PSNR/SSIM, as in prior works~\cite{bhat2021deep,bhat2021deep1,dudhane2022burst}.
            Instead of training from scratch, we fine-tune the pre-trained model (of SyntheticBurst dataset) for 100 epochs on the BurstSR dataset.
            Table~\ref{tab: burstSR} shows that our \xnet performs favorably well by providing PSNR gain of 0.33 dB over the previous best method BIPNet~\cite{dudhane2022burst}.
            We present visual comparisons in Fig.~\ref{fig: burstSR}. \xnet-generated images exhibit higher detail, sharpness, and visual accuracy.
         
\begin{figure}[t]
            \centering
            \includegraphics[width=1\linewidth]{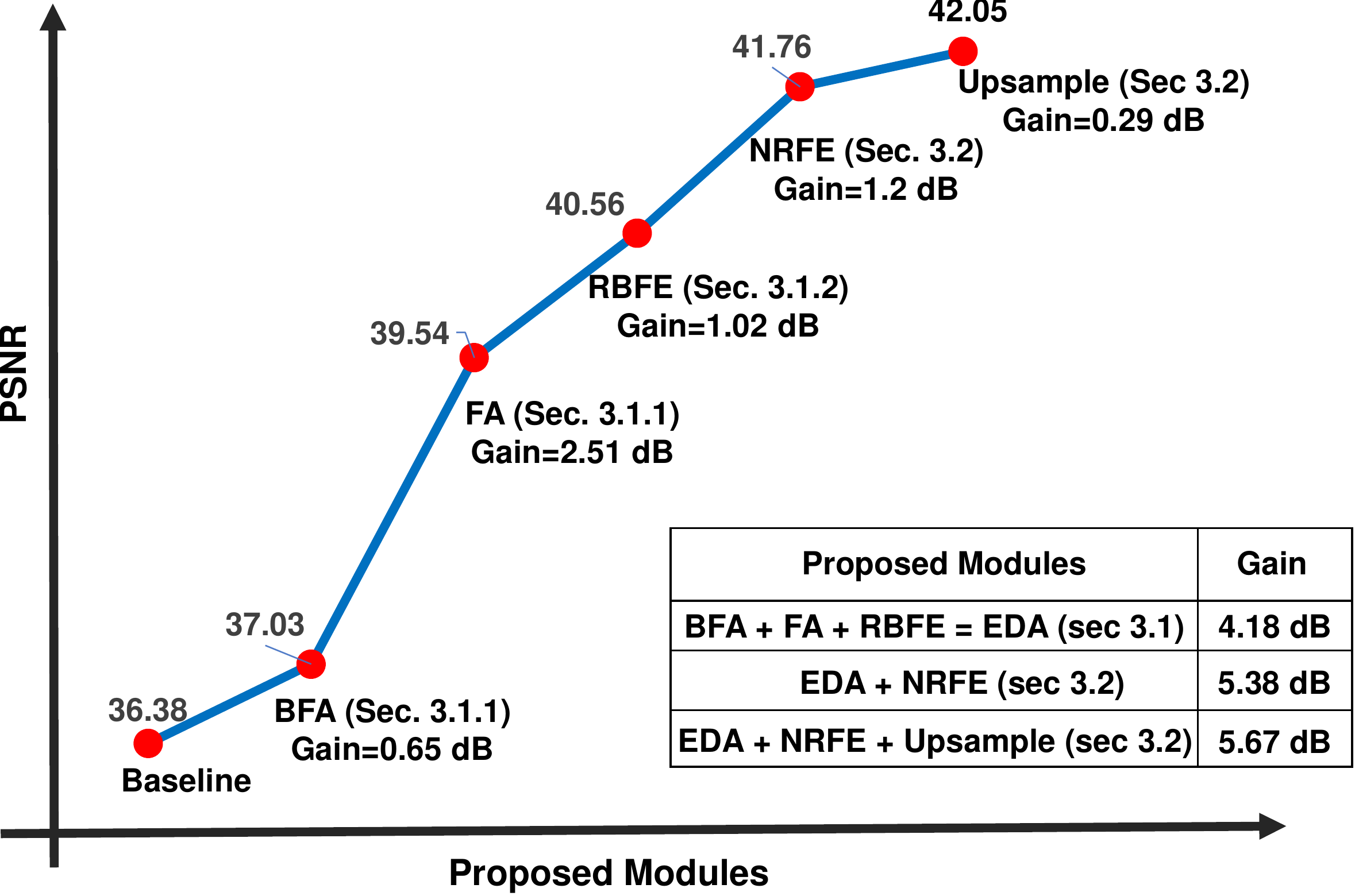}
            \caption{\textbf{Ablation experiments} for \xnet contributions. PSNR is reported on SyntheticBurst dataset~\cite{bhat2021deep} for $4 \times$ SR. All our major components contribute significantly. As given in Table, our \xnet achieves 5.67 dB gain over the baseline approach.}
            \label{fig: ablations1}
        \end{figure}

        \vspace{0.4em}
        \noindent \textbf{Ablation experiments.}
            To study the impact of different modules of the proposed architecture on the final performance, we train several ablation models on the SyntheticBurst dataset~\cite{bhat2021deep} for 100 epochs. Results are provided in Fig.~\ref{fig: ablations1}.
            In the baseline model, we use Resblocks~\cite{lim2017edsr} for feature extraction, simple concatenation-based fusion, and the pixel-shuffle operation for upsampling. 
            It can be seen that inclusion of the proposed RBFE in feature alignment stage leads to substantial PSNR boost of 1.02 dB. 
            This performance gain is further increased by 1.49 dB when we add the proposed burst fusion (NRFE) and upsampling modules.
            Overall, when deployed all our modules, we achieve 5.67 dB increment over the baseline.
            Further, Table~\ref{tab: ablation2} shows that replacing the proposed alignment and fusion methods with other existing techniques causes significant performance drop, \textit{i.e.,} 0.43 dB and 0.34 dB, respectively.
            Specifically, our \xnet lead to 0.79 dB boost when compared with existing multi-level PCD alignment~\cite{wang2019edvr}. The proposed RBFE module with local-non-local feature extraction ability is a key difference between the existing PCD and our enhanced deformable alignment. Further, we observe 0.34 dB drop in PSNR when we replace the proposed NRFE (fusion module) with existing compute extensive PBFF~\cite{dudhane2022burst}. Ablation experiments show that with compute efficient in nature our modules outperform other existing modules in all manner without any compromise in performance. 

        \begin{table}[t]
            \caption{Comparison of \textbf{alignment and fusion} techniques. PSNR is computed on SyntheticBurst~\cite{bhat2021deep} for $4 \times$ SR.}
            \label{tab: ablation2}
            \centering
            \setlength{\tabcolsep}{28pt}
            \scalebox{0.8}{
                \begin{tabular}{l|l@{$\;\,$}|l@{$\;\,$}}
                    \toprule
                      \textbf{Task}  & \textbf{Methods}  & \textbf{PSNR $\uparrow$} \\
                    \midrule
                    \multirow{3}[0]{*}{\textbf{Alignment}} & Explicit~\cite{bhat2021deep} & 39.84 \\
                    & TDAN~\cite{tian2020tdan} & 40.58 \\
                    & PCD~\cite{wang2019edvr} & 41.26 \\
                    & EBFA~\cite{dudhane2022burst} & 41.62 \\
                    & \textbf{\xnet~(Ours)} & \textbf{42.05} \\
                    \midrule
                    \multirow{3}[0]{*}{\textbf{Burst Fusion}} & Addition & 40.20 \\
                    & Concat & 40.65 \\
                    & DBSR~\cite{bhat2021deep} & 41.08 \\
                    & PBFF~\cite{dudhane2022burst} & 41.71 \\
                    \midrule
                    & \textbf{\xnet~(Ours)} & \textbf{42.05} \\
                    \bottomrule
                \end{tabular}
                }
        \end{table}
         
    \subsection{Burst Low-Light Image Enhancement}
        We test the performance of our \xnet on the Sony subset from the SID dataset, as in other existing works~\cite{dudhane2022burst,zamir2020learning,karadeniz2020burst,zhao2019end}.
        In addition to $L_1$ loss, we use the perceptual loss~\cite{zhang2018unreasonable} for network optimization.
        
        \vspace{0.4em}
        \noindent \textbf{Dataset.} 
            SID~\cite{chen2018learning} contains input RAW burst sequences captured with short-camera exposure in extreme low ambient light, and their corresponding well-exposed sRGB ground-truth images. The dataset consists of 161 burst sequences for training, 36 for validation, and 93 for testing.
            We crop 6,500 patches of size $256$$\times$$256$ with burst size varying from 4 to 8 and train the network for 200 epochs. Since the input RAW burst is mosaicked, we use single $2 \times$ upsampler in our \xnet to obtain the final image.

        \begin{table}
            \caption{\textbf{Burst low-light image enhancement} evaluation on the SID dataset \cite{chen2018learning}. Burstromer performs well across three metrics.}
            \label{tab: enhancement}
            \centering
            \setlength{\tabcolsep}{13pt}
            \scalebox{0.8}{
                \begin{tabular}{l | c | c | c}
                    \toprule
                    \textbf{Methods} & \textbf{PSNR $\uparrow$} & \textbf{SSIM $\uparrow$} & \textbf{LPIPS $\downarrow$} \\
                    \midrule
                    Chen \textit{et al.}\cite{chen2018learning} & 29.38 & 0.892 & 0.484 \\
                    Maharjan \textit{et al.} \cite{maharjan2019improving} & 29.57 & 0.891 & 0.484 \\
                    Zamir \textit{et al.} \cite{zamir2021learning} & 29.13 & 0.881 & 0.462 \\
                    Zhao \textit{et al.} \cite{zhao2019end} & 29.49 & 0.895 & 0.455 \\
                    LEED \cite{karadeniz2020burst} & 29.80 & 0.891 & 0.306 \\
                    BIPNet  \cite{dudhane2022burst} & 32.87 & 0.936 & 0.305 \\
                    \midrule
                    \textbf{Ours} & \textbf{33.34} & \textbf{0.941} & \textbf{0.285} \\
                    \bottomrule
                \end{tabular}
                }                      
        \end{table}

        \begin{figure}[t]
            \centering
            \vspace{1.5em}
            \includegraphics[width=1\linewidth]{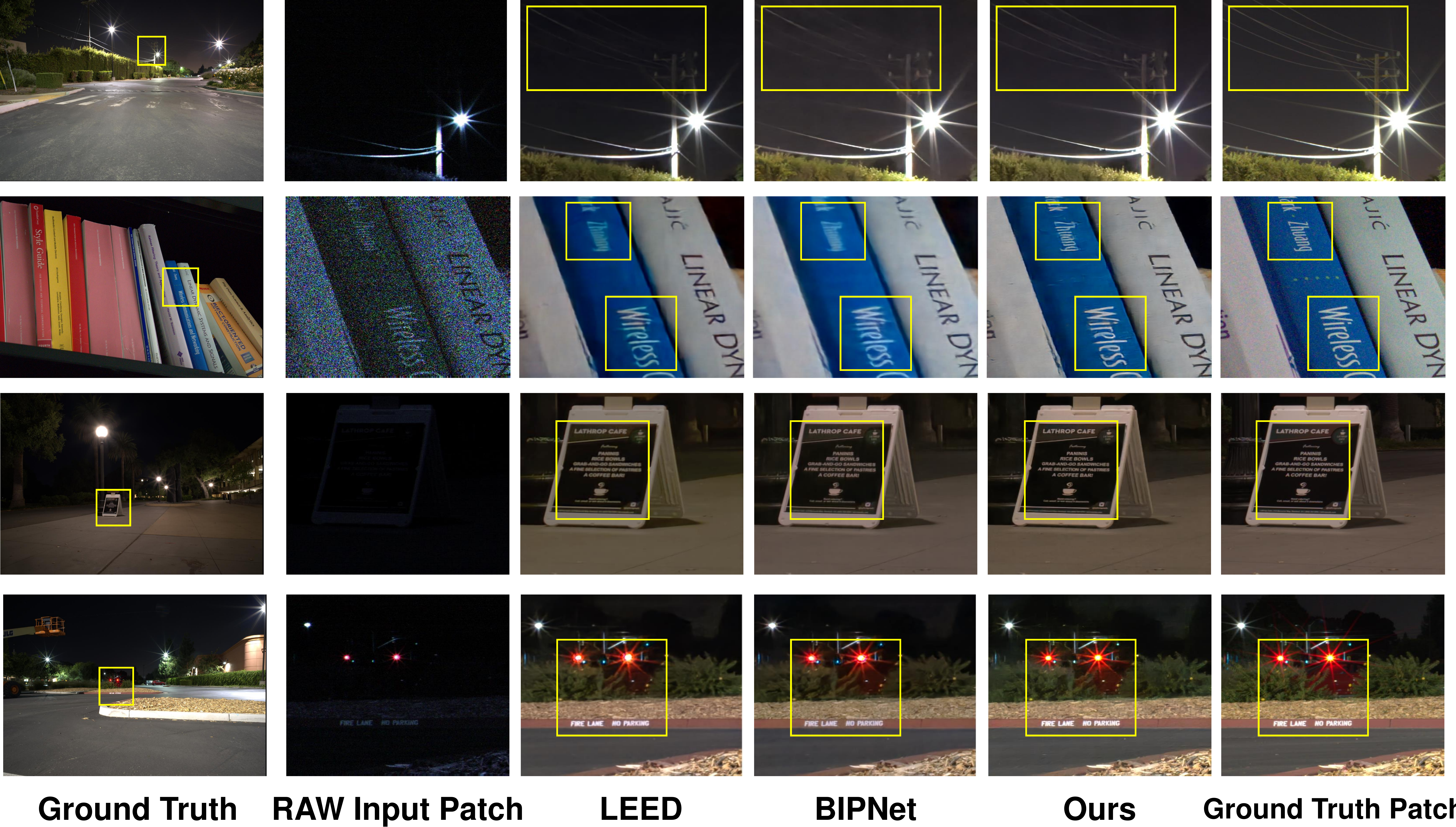}
            \caption{\textbf{Burst low-light image enhancement} comparisons on the Sony subset of SID dataset~\cite{chen2018learning}. Our \xnet retains color and structural details faithfully relative to the ground-truth.}
            \label{fig: enh results}
             \vspace{0.5em}
        \end{figure}
        
        \begin{figure*}[t]
            \centering
            \includegraphics[width=0.994\linewidth]{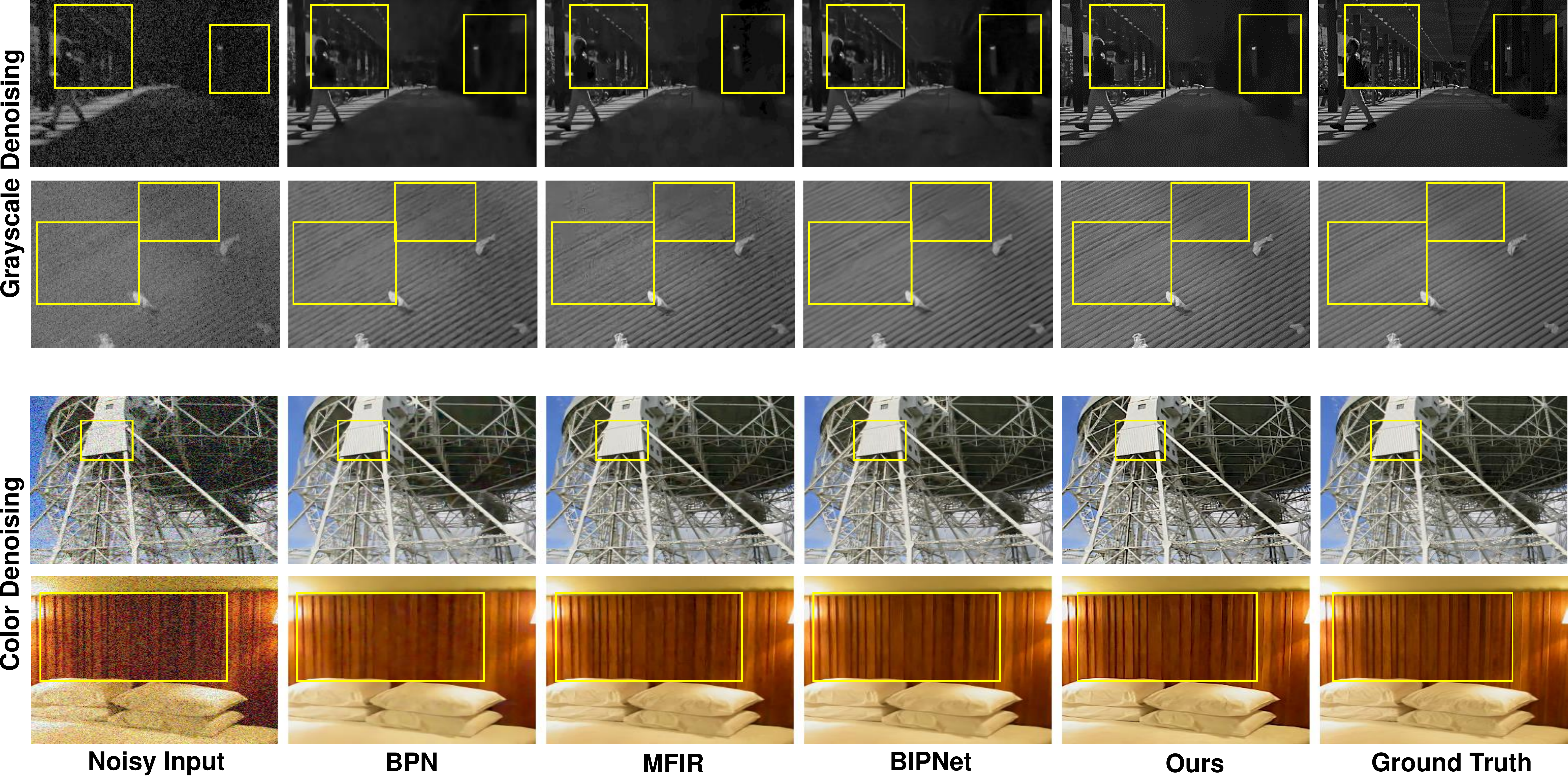}
            \vspace{0.07cm}
            \caption{\textbf{Burst denoising} results on burst images from the grayscale~\cite{Mildenhall2018BurstDW} and color datasets~\cite{Xia2020BasisPN}. Our \xnet produces more sharp and clean results than other competing approaches. More examples are provided in the supplementary material.}
            \label{fig: col den results}            
        \end{figure*}

        \vspace{0.4em}
        \noindent \textbf{Enhancement results.}
            The image quality scores for competing approaches are summarized in Table~\ref{tab: enhancement}. 
            Our \xnet achieves PSNR gains of 0.47 dB over the previous best method BIPNet~\cite{dudhane2022burst} and 3.54 dB over the second best algorithm LEED~\cite{karadeniz2020burst}.
            Figure~\ref{fig: enh results} shows enhanced images produced by different approaches. Our \xnet yields images with more faithful color and structural content than the other compared approaches. 

    \subsection{Burst Denoising}
        This section presents the results of burst denoising on grayscale data~\cite{Mildenhall2018BurstDW} as well as on color data~\cite{Mildenhall2018BurstDW}.
        As there is no need to upscale the burst features, we replace the upsampler in \xnet with a simple convolution to generate the output image.
        
        \vspace{0.4em}
        \noindent \textbf{Datasets.}
            Following the experimental protocols of \cite{Mildenhall2018BurstDW} and \cite{Xia2020BasisPN}, we prepare training datasets for grayscale denoising and color denoising, respectively. We train separate denoising models for 300 epochs on 20K synthetic burst patches. Each burst contains 8 frames of 128$\times$128 spatial resolution.
            Testing is performed on 73 grayscale bursts and 100 color bursts. %
            Both of these test sets contain 4 variants with different noise gains $(\text{1,2,4,8})$, corresponding to noise parameters $(\log(\sigma_r),\log(\sigma_s)) \rightarrow$ $ (\text{-2.2,-2.6})$, $(\text{-1.8,-2.2})$, $(\text{-1.4,-1.8})$, and $(\text{-1.1,-1.5})$, respectively.
        
        \vspace{0.4em}
        \noindent \textbf{Denoising results.}
            We compare various existing methods such as KPN~\cite{Mildenhall2018BurstDW}, MKPN, BPN~\cite{Xia2020BasisPN}, MFIR~\cite{bhat2021deep1}, and BIPNet~\cite{dudhane2022burst}.
            Since the proposed \xnet trained without any extra data or supervision, we consider results of the  MFIR~\cite{bhat2021deep1} variant that uses a custom optical flow sub-network (without pre-training it on extra data).
            Table~\ref{tab:kpn_grayscale} reports grayscale denoising results where our \xnet consistently performs well. When averaged across all noise levels, our method provides 0.75 dB PSNR improvement over the state-of-the-art BIPNet~\cite{dudhane2022burst}\footnote[2]{\label{note}We use BIPNet results from the official Github repository.}.
            Table~\ref{tab:kpn_color} shows that the performance trend of \xnet is similar on color denoising as well. For instance, on high noise level bursts (Gain $\propto$ 8), \xnet provides PSNR boost of 0.57 dB over BIPNet~\cite{dudhane2022burst}. 
            Visual comparisons in Fig.~\ref{fig: col den results} show that  \xnet's denoised outputs are relatively cleaner, sharper and preserve subtle textures. Additional qualitative results are provided in supplementary material.

        \begin{table}[t]
            \caption{\textbf{Grayscale burst denoising} on the  dataset by \cite{Mildenhall2018BurstDW}. PSNR is reported.}
            \label{tab:kpn_grayscale}
            \centering
            \setlength{\tabcolsep}{2.3pt}
            \scalebox{0.8}{
                \begin{tabular}{l|cccc|c}
                    \toprule
                    & Gain {$\propto$} 1 & Gain {$\propto$} 2 & Gain {$\propto$} 4 & Gain {$\propto$} 8 & \textbf{Average}\\
                    \midrule                
                    KPN~\cite{Mildenhall2018BurstDW}&36.47&33.93&31.19&27.97&32.39\\
                    BPN~\cite{Xia2020BasisPN}&38.18&35.42&32.54& 29.45 & 33.90\\
                    MFIR~\cite{bhat2021deep1} & 39.10 & 36.14 & 32.89 & 28.98 & 34.28 \\
                    BIPNet~\cite{dudhane2022burst}$^{\ref{note}}$ & 38.53 & 35.94 & 33.08 & 29.89 & 34.36 \\
                    \midrule
                    \textbf{\xnet~(Ours)} & \textbf{39.49} & \textbf{36.70} & \textbf{33.71} & \textbf{30.55} & \textbf{35.11} \\
                    \bottomrule
                \end{tabular}}
        \end{table}
                
        \begin{table}[t]
            \caption{\textbf{Color burst denoising} on the  dataset by \cite{Xia2020BasisPN}. PSNR is reported.}
            \label{tab:kpn_color}%
            \centering
            \setlength{\tabcolsep}{2.3pt}
            \scalebox{0.80}{
                \begin{tabular}{l|cccc|c}
                    \toprule
                    &Gain $\propto$ 1 & Gain $\propto$ 2 & Gain $\propto$ 4 & Gain $\propto$ 8 & \textbf{Average} \\
                    \midrule                    
                    KPN~\cite{Mildenhall2018BurstDW} & 38.86 & 35.97 & 32.79 & 30.01 & 34.41 \\
                    BPN~\cite{Xia2020BasisPN} & 40.16 & 37.08 & 33.81 & 31.19 & 35.56 \\
                    BIPNet~\cite{dudhane2022burst}$^{\ref{note}}$ & 40.58 & 38.13 & 35.30 & 32.87 & 36.72\\
                    MFIR~\cite{bhat2021deep1} & \textbf{41.90} & 38.85 & 35.48 & 32.29 & 37.13 \\
                    \midrule
                    \textbf{\xnet~(Ours)} & 41.70 & \textbf{39.15} & \textbf{36.09} & \textbf{33.44} & \textbf{37.59} \\
                    \bottomrule
                \end{tabular}}
        \end{table}
           
\section{Conclusion}

    We present a transformer-based framework for burst image processing. The proposed \xnet is capable of generating a single high-quality image from a given burst of noisy images having pixel misalignments among them.
    \xnet employs a multi-scale hierarchical module EDA that, at each scale, first generates denoised features encoding local and non-local context, and then aligns each burst frame with the reference frame.
    To fix any remaining minor alignment issues, we incorporate a reference-based feature enrichment RBFE module in EDA that enables additional interaction of the features of each frame with the base frame features.
    Overall, EDA improves model robustness by yielding a burst of features that are well denoised, aligned, consolidated and refined. 
    In the image reconstruction stage, we repeatedly apply the no-reference feature enrichment NRFE and upsampling modules in tandem until the final image is obtained. 
    NRFE progressively and adaptively fuses each pair of frame features that are obtained with the proposed cyclic burst sampling.
    Experiments performed on three representative burst processing tasks (super-resolution, denoising, low-light image enhancement) demonstrate that our \xnet provides state-of-the-art results and generalizes well compared to recent burst processing approaches.

{\small
\bibliographystyle{ieee_fullname}
\bibliography{arxive}
}

\clearpage

\renewcommand\thesection{\Alph{section}}
\setcounter{section}{0}

\setcounter{table}{0}
\renewcommand{\thetable}{S\arabic{table}}

\setcounter{figure}{0}
\renewcommand{\thefigure}{S\arabic{figure}}

\onecolumn
\begin{center}
\textbf{\LARGE Supplemental Material}
\end{center}
\vspace{2cm}

    Here we provide more details on architectural design, additional ablations, and visual comparisons for burst SR, low-light image enhancement and denoising.
    
    \section{Network Architectural Details}
        In \xnet, the EDA module is a 3-level encoder-decoder, where each level employs 1 FA (containing single deformable conv. layer) and 1 RBFE module. In the image reconstruction stage, we use 2 NRFE modules. The BFF unit both in RBFE and NRFE consists of 1 BFA module. 
        
\begin{figure*}[ht]
        \centering
        \includegraphics[width=1\linewidth]{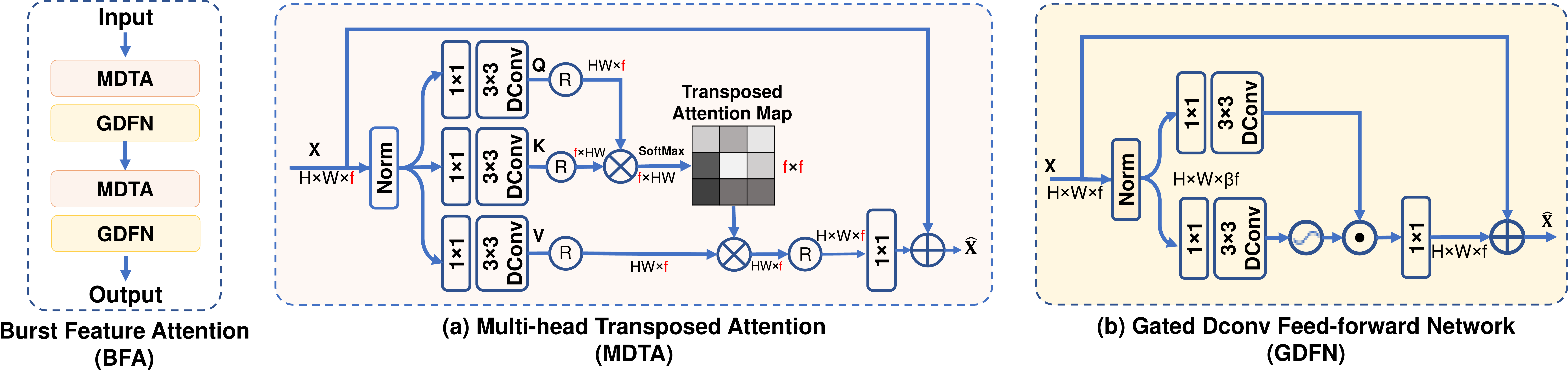}
        \caption{\small \textbf{Burst Feature Attention} (BFA) used in the proposed alignment and reconstruction stages to extract features encoding both local and non-local pixel interactions.}
        \label{fig: supp_BFA}
    \end{figure*}

        Figure \ref{fig: supp_BFA} shows the BFA module that consists of multi-dconv head transposed attention (MDTA) and gated-Dconv feed-forward network (GDFN)~\cite{zamir2021restormer}.
        MDTA encodes local and non-local context, and efficient enough to be applied to high-resolution images. 
        Whereas, GDFN performs controlled feature transformation i.e., suppressing less informative features, and allowing only the useful information to pass further through the network.
    
    \section{Ablations on alignment and fusion modules}    
        Table \ref{tab:align_comp} compares the the properties of the proposed EDA and other existing alignment modules.
        Unlike existing explicit feature alignment approaches DBSR~\cite{bhat2021deep} and MFIR~\cite{bhat2021deep1}, the proposed EDA operates at multiple spatial scales and aligns burst features implicitly without any additional supervision.
        Overall, the proposed EDA module possesses required properties which makes it effective for the burst feature alignment.
        
        Table \ref{tab:fusion_comp} compares several feature fusion techniques.
        Our NRFE is flexible to taking as input the features of more than two frames. It extracts local and non-local burst features, enables long-range inter-frame interactions and aggregates the burst neighborhoods to obtain high-quality image.
    
    \section{Additional visual results}
        \label{fig: results appendix}
\begin{figure*}[t]
        \centering
        \includegraphics[width=1\linewidth]{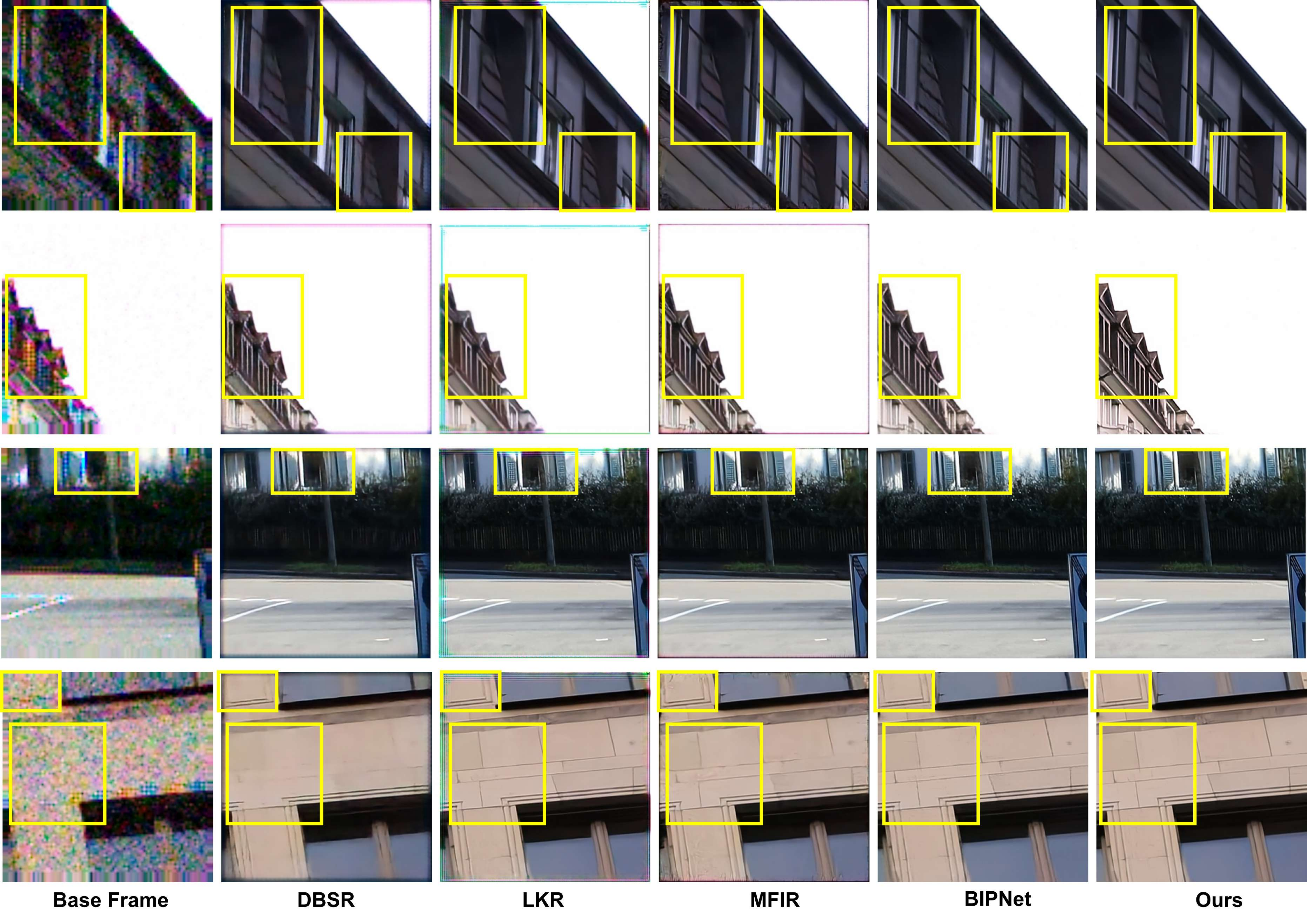}
        \caption{\small \textbf{Burst super-resolution} ($4 \times$) results on SyntheticBurst dataset~\cite{bhat2021deep}. }
        \label{fig: supp_syn_SR}
    \end{figure*}

\begin{figure*}[t]
        \centering
        \includegraphics[width=1\linewidth]{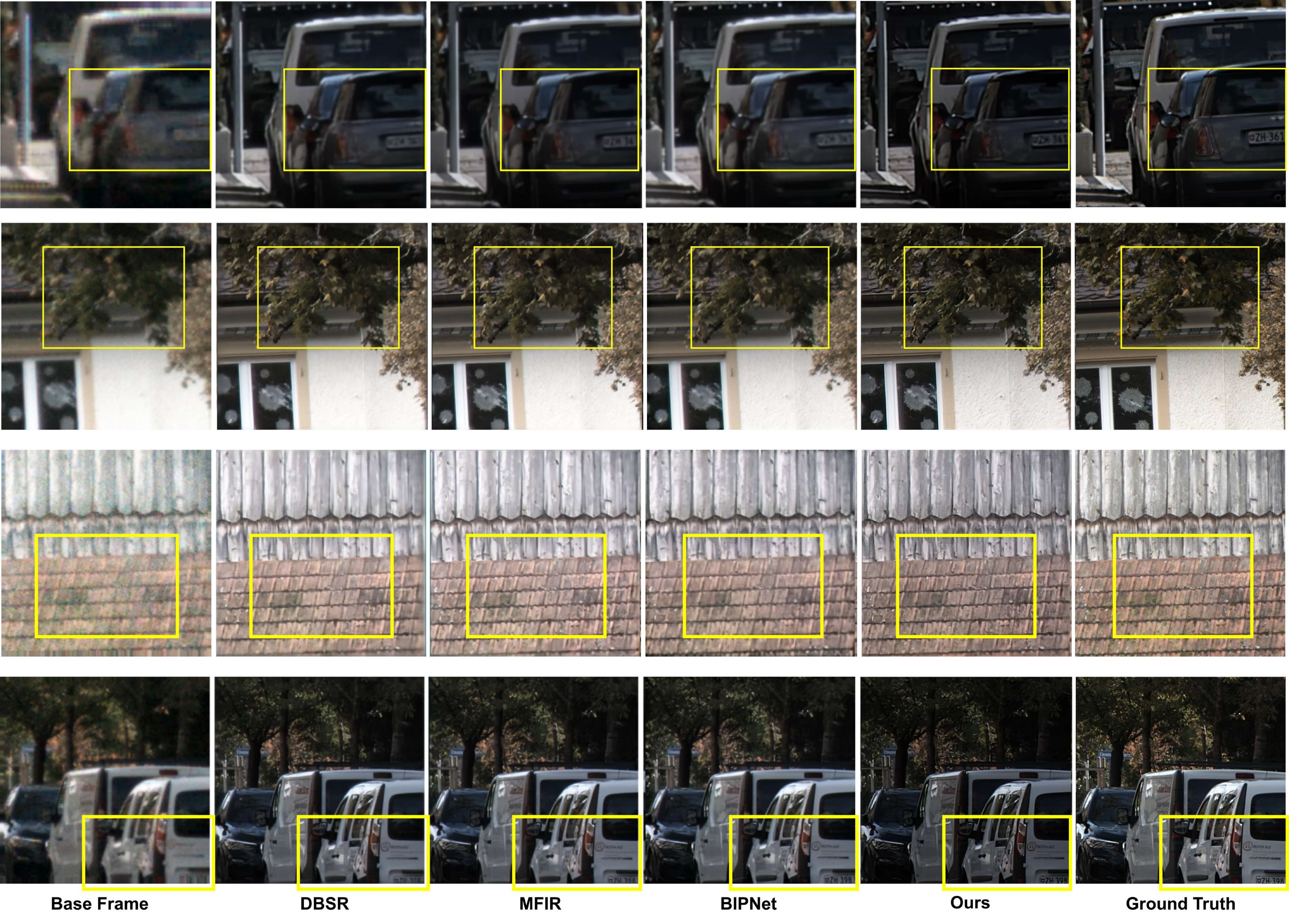}
        \caption{\small \textbf{Burst super-resolution} ($4 \times$) results on BurstSR (real) dataset~\cite{bhat2021deep}. }
        \label{fig: supp_real_SR}
    \end{figure*}

        \textbf{Burst Super-resolution.}    Figure~\ref{fig: supp_syn_SR}, and Figure~\ref{fig: supp_real_SR} show qualitative results of competing approaches on examples from the SyntheticBurst and (real) BurstSR datasets ~\cite{bhat2021deep} for $4\times$ SR. The reproductions of our \xnet are more detailed, sharper than those produced by the other methods.
            
\begin{figure*}[t]
        \centering
        \includegraphics[width=1\linewidth]{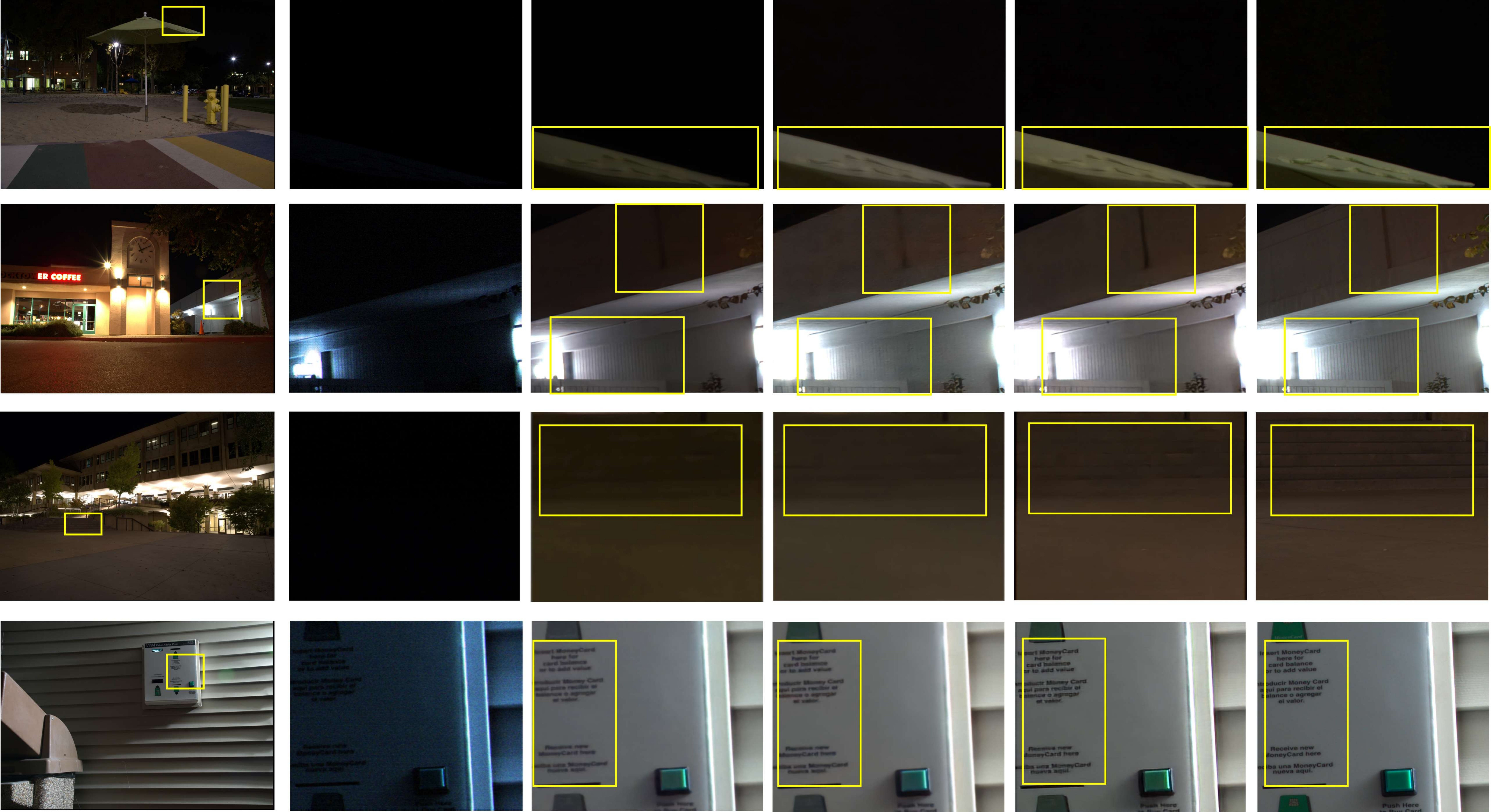}
        \caption{\textbf{Burst low-light image enhancement} comparisons on the Sony subset of SID dataset~\cite{chen2018learning}.}
        \label{fig: supp_enh}
    \end{figure*}

        \textbf{Burst low-light image enhancement.} Figure~\ref{fig: supp_enh} depicts that \xnet produces images that are visually more closer to the ground-truth than the other approaches.  
            
\begin{figure*}[t]
        \centering
        \includegraphics[width=1\linewidth]{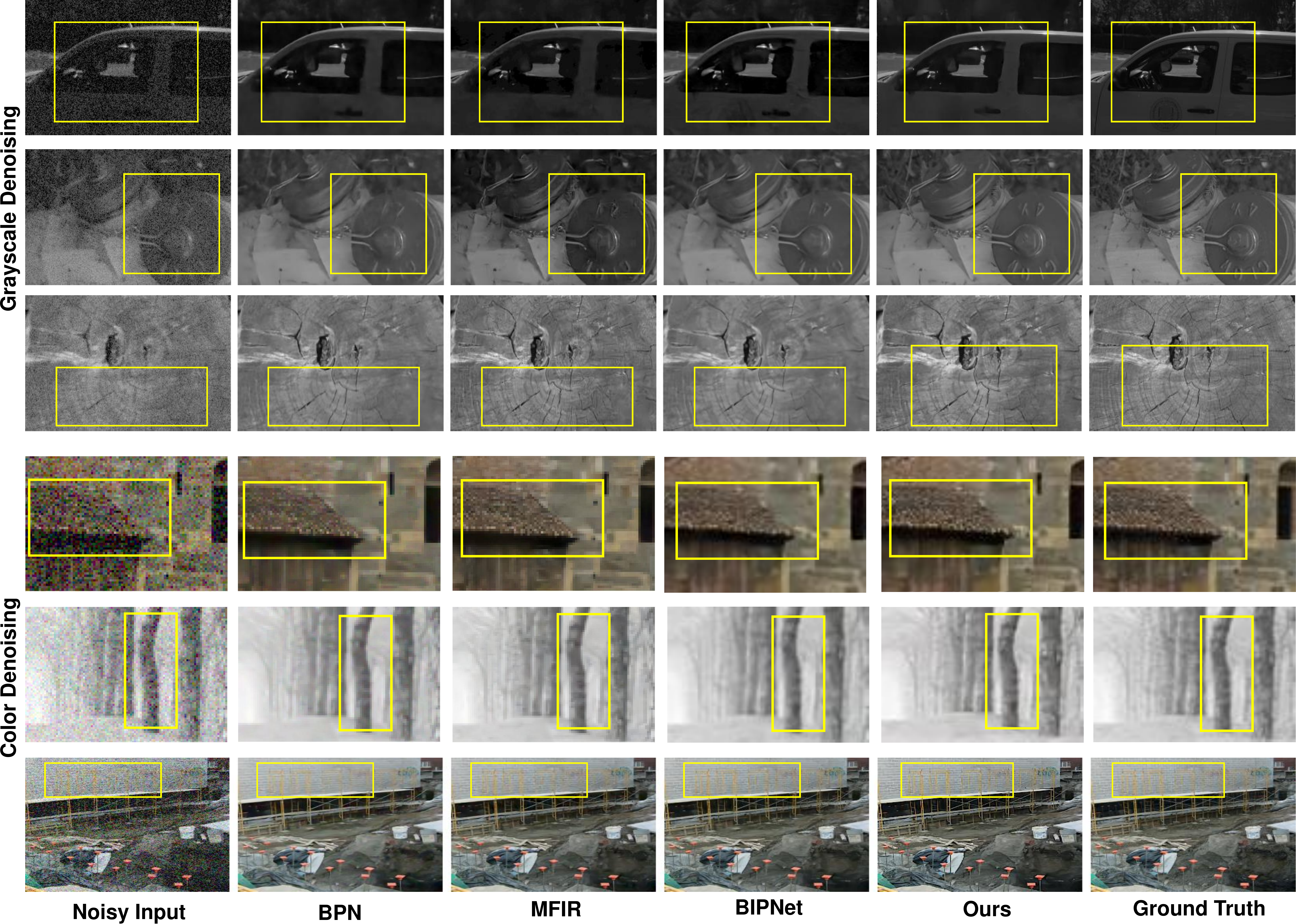}
        \caption{\textbf{Burst denoising} results on burst images from the grayscale~\cite{Mildenhall2018BurstDW} and color datasets~\cite{Xia2020BasisPN}.}
        \label{fig: supp_denoise}
    \end{figure*}

        \textbf{Burst Denoising.} Figure \ref{fig: supp_denoise} shows that the proposed \xnet is capable of removing noise, while preserving the desired texture and structural content. 
    \vspace{1cm}
    \begin{table*}[t]
        \setlength{\tabcolsep}{4pt}
        \centering
        \small        
        \caption{Ablation on existing \textbf{Feature alignment strategies} with our EDA module. }
        \begin{tabular}{l ccccc}
            \toprule
                  & \textbf{DBSR/MFIR}~\cite{bhat2021deep, bhat2021deep1} & \textbf{TDAN}~\cite{tian2020tdan} & \textbf{PCD}~\cite{wang2019edvr} & \textbf{EBFA}~\cite{dudhane2022burst} & \textbf{EDA (Ours)}\\
            \midrule
            Extra supervision & \ch & $\times$ & $\times$ & $\times$ & $\times$\\
            Implicit alignment & $\times$ & \ch & \ch & \ch & \ch\\
            Multi-scale hierarchy & \ch & $\times$ & \ch & $\times$ & \ch\\
            Attention for feature denoising & $\times$ & $\times$ & $\times$ & \ch & \ch\\
            Reference-frame based refinement & $\times$ & $\times$ & $\times$ & $\times$ & \ch \\            
        \bottomrule
        \end{tabular}%
        \label{tab:align_comp}%
    \end{table*}%
    \begin{table*}[t]
        \setlength{\tabcolsep}{4pt}
        \centering
        \small        
        \caption{Ablation on existing \textbf{Feature fusion techniques} with our NRFE module. }
        \begin{tabular}{l ccc}
            \toprule
            & \textbf{DBSR/MFIR}~\cite{bhat2021deep, bhat2021deep1} & \textbf{PBFF}~\cite{dudhane2022burst} & \textbf{NRFE (Ours)}\\
            \midrule
            Flexible w.r.t multiple inputs & \ch & $\times$ & \ch\\
            Long-range inter-frame interaction & $\times$ & \ch & \ch\\
            Local and non-local feature extraction & $\times$ & \ch & \ch\\
            Computational overhead & $\downarrow$ & $\uparrow$ & $\downarrow$\\
            \bottomrule
        \end{tabular}%
        \label{tab:fusion_comp}%
    \end{table*}%

\end{document}